\documentclass[letterpaper]{article} % DO NOT CHANGE THIS
\usepackage{aaai2027}  % DO NOT CHANGE THIS
\usepackage[hyphens]{url}  % DO NOT CHANGE THIS
\usepackage{graphicx} % DO NOT CHANGE THIS
\usepackage{natbib}  % DO NOT CHANGE THIS AND DO NOT ADD ANY OPTIONS TO IT
\usepackage{caption} % DO NOT CHANGE THIS AND DO NOT ADD ANY OPTIONS TO IT
\usepackage{multirow, makecell}

\usepackage{algorithm}
\usepackage{algorithmic}

\usepackage{amsmath}        % align, equation environments
\usepackage{amssymb}        % \mathbb{R}
\usepackage{bm}             % bold math (optional but useful)

\usepackage{newfloat}
\usepackage{listings}

\DeclareCaptionStyle{ruled}{labelfont=normalfont,labelsep=colon,strut=off} % DO NOT CHANGE THIS
\floatstyle{ruled}
\newfloat{listing}{tb}{lst}{}
\floatname{listing}{Listing}

\usepackage{booktabs}

\title{CIIR — Class-Indexed Interface Representation Learning}

\author{
    Doo Yon Kim,\\
    Christian Wallraven
}
\affiliations{
    \textsuperscript{\rm 1}Department of Artificial Intelligence, Korea University, Seoul, Korea\\
   \textsuperscript{\rm 2} Brain and Cognitive Engineering and Department of Artificial Intelligence, Korea University, Seoul, Korea\\
    
    kdoodoo@korea.ac.kr, wallraven@korea.ac.kr 
}

\begin{document}

\maketitle

\begin{abstract}
Multimodal learning commonly assumes paired or synchronized observations, yet independently collected datasets rarely provide such correspondence. We introduce CIIR, a modular encoder--decoder framework that connects heterogeneous datasets through a class-indexed address space while preserving each dataset's native input format and output module. A known local index selects address $k$, and an input-dependent 128-dimensional address representation $Z_k(x)$ is written to that address. To demonstrate CIIR, we learn this address space from handwritten digits, clothing images, spoken-digit spectrograms, and hand signs without synchronized or instance-matched samples. Newly trained label-free receivers recover the requested addresses with 94.7\% macro-average Address Realization Accuracy. We then attach the unseen SVHN dataset after freezing the base address space and train only lightweight adapters, achieving 90.2\% MNIST$\rightarrow$SVHN and 82.3\% SVHN$\rightarrow$MNIST end-to-end Address Transfer Accuracy without target-side decoder labels. To demonstrate generalizability of CIIR, we reuse the ten addresses as code elements in collision-free arithmetic and hierarchical address codes for all 100 CIFAR-100 classes and evaluate direct arithmetic-code realization through the existing receivers. These results establish CIIR as a compact interface for modular cross-dataset communication and post-hoc dataset attachment without exhaustive paired multimodal supervision.
\end{abstract}

\section{Introduction}

Multimodal learning commonly assumes paired or synchronized observations. This assumption is appropriate when several sensors record the same event, but restrictive when independently collected datasets and modality-specific models must be integrated after training. Requiring each new dataset to be paired with every existing modality limits modular reuse and increases the cost of extending a multimodal system. Most multimodal representation-learning methods address missing modalities by learning shared latent variables from paired or co-occurring observations. Multimodal variational autoencoders combine modality-specific posteriors through products, mixtures, or related aggregation rules to support cross-generation and incomplete-input inference \cite{suzuki2017jmvae,wu2018mvae,shi2019mmvae,sutter2021mopoe}. Later work improves shared--private factorization, information-theoretic objectives, posterior modeling, and generative quality \cite{lee2021dmvae,hwang2021mvtcae,palumbo2023mmvaeplus, senellart2025inferencegap,cui2026shala,vo2026helvae,vo2026holderpp}. These methods ask how corresponding modalities can be represented jointly; we instead ask whether independently collected datasets can communicate through a reusable interface without instance-level correspondence.

We introduce \textbf{CIIR}, \textbf{Class-Indexed Interface Representation Learning}, a modular representation framework that learns input-dependent address representations in a shared address space and supports post-hoc connection of independently collected datasets without retraining previously learned components. In CIIR, the address identifies the class index, while the address
representation carries input-specific information. A known local class index $k$ selects address $k$, while $Z_k(x)$ carries the transmitted information. Importantly, equal addresses need not share semantics across datasets: address $k$ may denote a digit in MNIST \cite{lecun2010mnist}, a clothing category image in Fashion-MNIST \cite{xiao2017}, a spoken-digit spectrogram in Free Spoken Digit Dataset (FSDD) \cite{fsdd}, or a remapped hand sign image in Sign Language MNIST~\cite{tecperson2017signmnist}. CIIR preserves dataset-specific output modules and trains a unified encoder with ten addresses, each carrying a 128-dimensional representation. Inputs with various native resolutions meet only in this fixed-dimensional address space. Unlike a fixed lookup table, CIIR produces input-dependent address representations that receivers decode into images or spectrograms rather than direct class predictions.

We evaluate CIIR in three stages. First, we test whether label-free receivers can recover the requested addresses across four independently collected datasets. Second, we freeze the learned address space and attach a new dataset, SVHN \cite{netzer2011svhn}, through lightweight read/write adapters to evaluate bidirectional transfer. Third, we show how the same ten addresses can be used as a discrete vocabulary for larger class spaces through arithmetic and hierarchical CIFAR-100 \cite{krizhevsky2009cifar} address codes.

CIIR's contributions are: (1) a reusable shared address space across independently collected datasets; (2) modular generation in each dataset's native output format without paired samples; (3) controlled evidence that addresses are realized through input-dependent representations rather than fixed class codes; and (4) post-hoc reuse of the frozen address space for new datasets and larger class spaces.

\section{Related Work}

\subsection{Multimodal Representation Learning}
Multimodal representation learning seeks shared information while retaining modality-specific structure \cite{baltrusaitis2019multimodal,andrew2013dcca,wang2015dccae}. Multimodal VAEs, for example, support missing-modality inference and cross-generation through joint or aggregated latent variables \cite{wu2018mvae,shi2019mmvae,sutter2020mmjsd,oshima2026enhancing,sutter2021mopoe}. Subsequent methods improve shared--private factorization, total-correlation objectives, contrastive alignment, posterior modeling, and the coherence--quality trade-off \cite{lee2021dmvae,hwang2021mvtcae,poklukar2022gmc, palumbo2023mmvaeplus,senellart2025inferencegap,cui2026shala, vo2026helvae,vo2026holderpp}. Recent aggregation methods include data-dependent priors, barycentric formulations, and dependent-expert models \cite{sutter2024unity,qiu2025barycentric,mancisidor2025code}. These approaches generally learn shared semantic structure from corresponding modalities. CIIR instead shares an address whose local meaning may differ across independently collected datasets.

\subsection{Unpaired and Extensible Multimodal Learning}

Missing-modality methods usually retain paired or partially paired training observations. Other work relaxes this requirement through limited coupled data, overlapping modalities, or independently pretrained representation spaces \cite{zhang2023unseen,norelli2023asif,wang2023cmcr, zhang2024exmcr,zhu2024freebind,wei2024omnibind}. For example, ASIF connects pretrained unimodal models using limited paired data \cite{norelli2023asif}, whereas C-MCR and Ex-MCR extend existing multimodal spaces through overlapping or previously aligned modalities \cite{wang2023cmcr,zhang2024exmcr}. Propensity-based alignment, relative representations, and latent-space translation infer correspondences or mappings between independently learned spaces \cite{xi2024propensity,moschella2023relative,maiorca2023latent}. These methods aim to recover or preserve semantic alignment. CIIR in contrast does not attempt to discover semantic correspondence from raw cross-dataset observations. Instead, each dataset's known local class index is assigned to address $k$ by design. Equal addresses may have different dataset-local meanings; for example, the same address may denote a handwritten digit, a clothing category, a spoken digit, or a hand sign. Some datasets nevertheless happen to share coarse concepts, such as digit identity in MNIST and FSDD. CIIR neither infers this overlap from paired examples nor requires it for learning the address space. Its contribution is therefore class-indexed representation learning without instance-level pairing, rather than learned semantic alignment across modalities.

This setting also differs from domain adaptation, which typically assumes a shared prediction task or label space across source and target domains \cite{ganin2016dann,long2018cdan,wang2018deep}. In our later MNIST--SVHN attachment experiment, shared digit identity is used deliberately to evaluate post-hoc transfer after the base address space has been frozen.

\subsection{Modular and Composable Representations}
Model stitching, representation matching, and latent translation connect separately trained components through lightweight mappings \cite{lenc2015understanding,csiszarik2021similarity, bansal2021stitching,moschella2023relative,maiorca2023latent}, although successful stitching need not imply equivalent encoded information \cite{smith2025functional}. Compositional and object-centric approaches instead seek reusable factors, entities, or slots that can be recombined beyond observed configurations \cite{bengio2013representation,higgins2017betavae, locatello2019challenging,burgess2019monet,greff2019iodine, locatello2020slotattention,engelcke2020genesis,he2021corl}. In this context, CIIR learns input-dependent address representations under an imposed address assignment and tests whether independently defined modules can read and write those representations. Fixed label codes, learned class embeddings, and address prototypes can support routing, but contain no input-dependent variation. CIIR therefore separates address $k$ from the address representation $Z_k(x)$ and tests the latter using label-free receivers, representation substitutions, address prototypes, and learned fixed address-vector baselines. Our later CIFAR-100 study uses addresses in predefined codes without learned semantic composition.

\begin{figure}[t]
\centering
\fontsize{7}{8}\selectfont
\includegraphics[width=\columnwidth]
{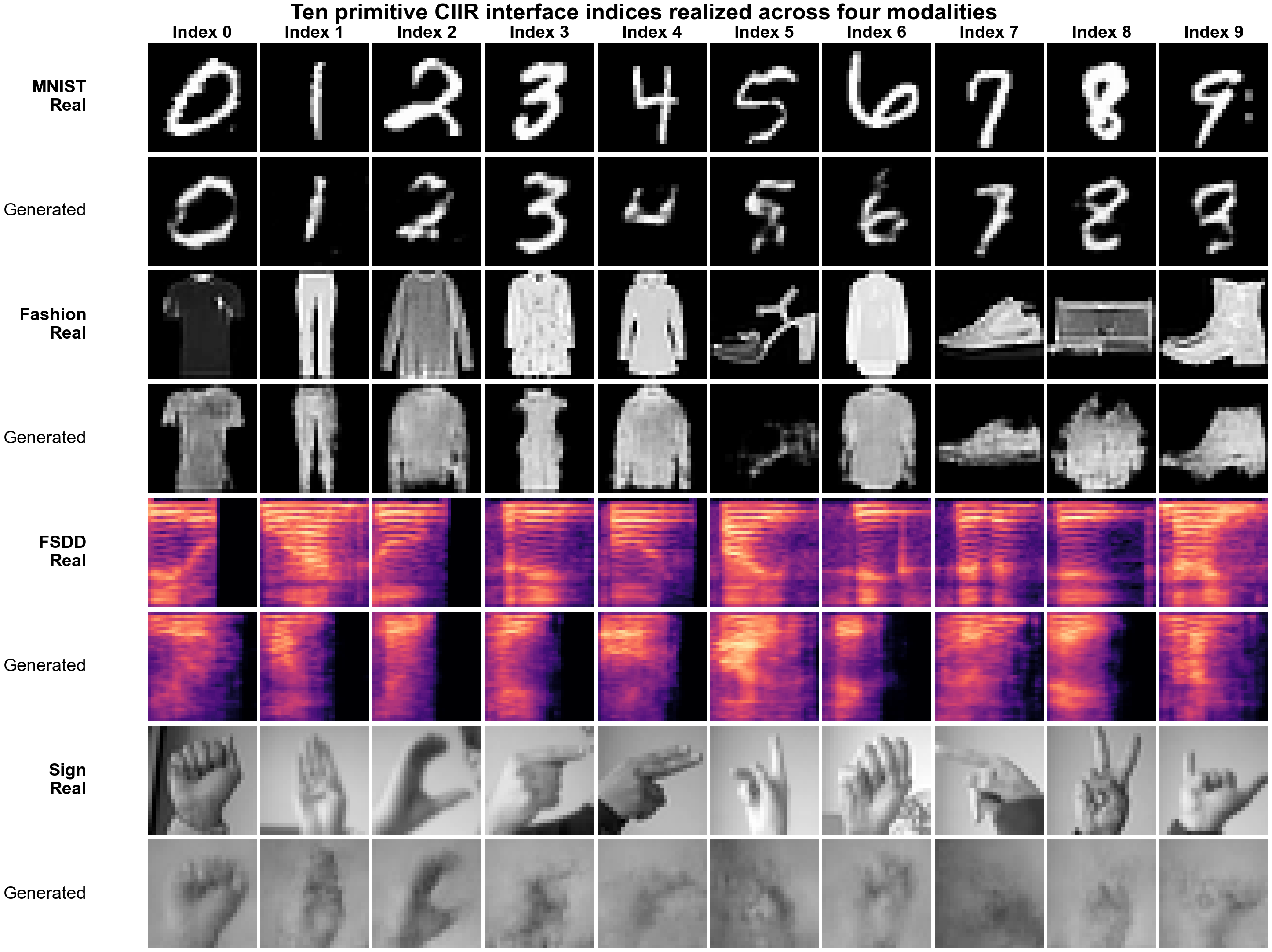}
\caption{
Qualitative realization of the ten CIIR addresses via their original label-conditioned decoders. Each pair of rows shows ground truth and generated outputs. Equal local indices select shared addresses while preserving dataset-local classes. For independent label-free evidence, see Table~\ref{tab:label_free_representation}.
}
\label{fig:base10_all_modalities}
\end{figure}

% ============================================================
% MAIN FIGURE 2 — ARITHMETIC AND HIERARCHICAL CIFAR-100 CODES
% ============================================================

\begin{figure}[ht]
\centering
\fontsize{7}{8}\selectfont

\includegraphics[width=\columnwidth]
{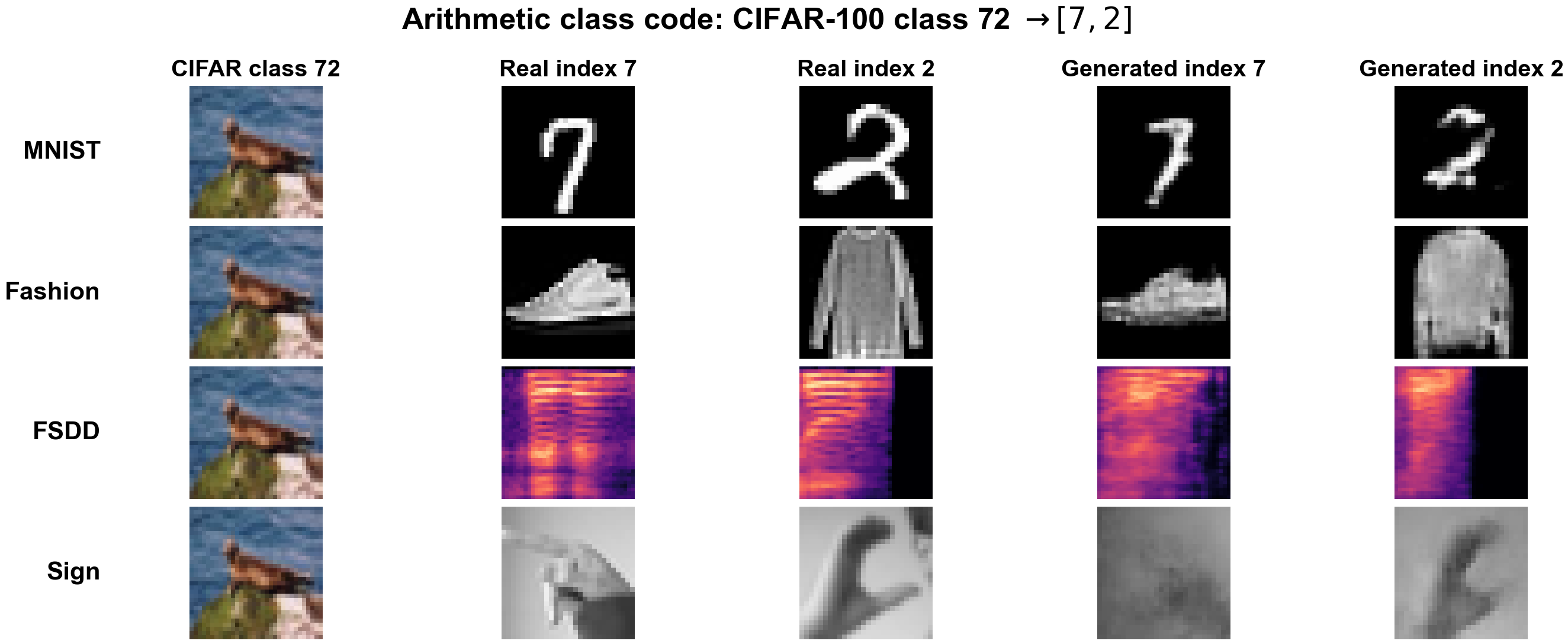}

\vspace{0.5mm}

\textbf{(a) Arithmetic code.}
Fine class $72$ is represented as
$\phi_{\mathrm{arith}}(72)=[7,2]$.

\vspace{2mm}

\includegraphics[width=\columnwidth]
{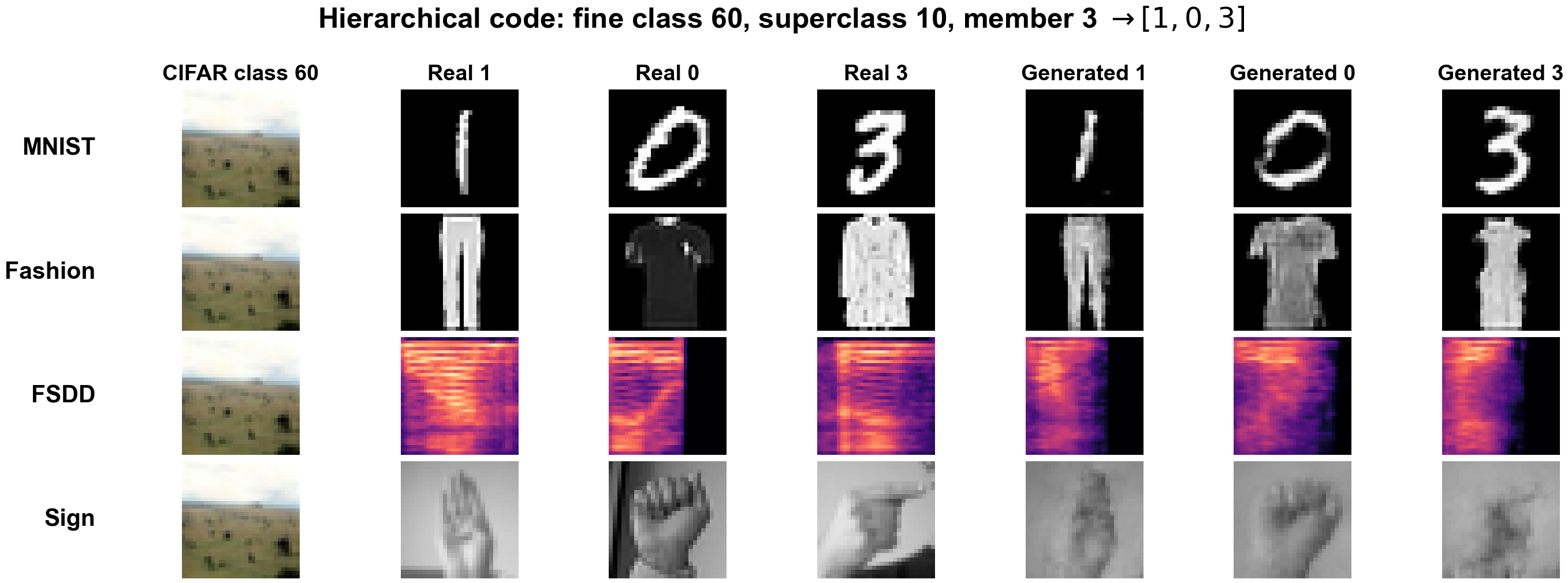}

\vspace{0.5mm}

\textbf{(b) Hierarchical code.}
A class in superclass $10$ with within-superclass member index $3$
is represented as $[1,0,3]$.

\caption{
Compositional application of CIIR to class-count mismatch across independently collected modalities. The ten learned addresses are reused as a coding vocabulary for the 100 CIFAR-100 classes: a known fine class is mapped to (a) the arithmetic address code $[7,2]$ or (b) the hierarchical address code $[1,0,3]$, and each requested address is realized in the native output domain of each modality.
}
\label{fig:cifar_code_realizations}
\end{figure}

\section{Method}
\label{sec:method}

Unlike multimodal approaches that learn a shared semantic representation from corresponding observations, CIIR first preserves each dataset's native output domain through a locally conditioned decoder and then connects these dataset-specific modules through a unified class-indexed address space. The method proceeds in three stages: (i) dataset-specific decoders are trained in their native output domains using local class indices; (ii) a unified encoder learns input-dependent representations organized by ten shared addresses; and (iii) the frozen address space is reused for post-hoc SVHN attachment and CIFAR-100 address coding. Equal address indices provide a common interface but need not denote the same class across datasets.

We use four representation terms throughout. An \emph{address} $k\in\{0,\ldots,K-1\}$ is a discrete location in the shared address space. An \emph{address representation} $Z_k(x)\in\mathbb{R}^{d}$ is the input-dependent vector written to address $k$. An \emph{address prototype} $p_k$ is the mean address representation associated with address $k$. An ordered sequence of addresses, such as $[1,9]$, forms an \emph{address code} for a larger class space.

We define five task-specific evaluation metrics. \emph{Address Realization Accuracy} (ARA) measures whether a receiver output is recognized as the requested address. \emph{Address Routing Accuracy} measures whether a source observation is assigned to its correct address. \emph{End-to-end Address Transfer Accuracy} measures whether the correct address is preserved through the complete source-to-address-to-target path. \emph{Transfer Retention} is the ratio of end-to-end transfer accuracy to Address Routing Accuracy. For multi-address codes, \emph{Class-Code Realization Accuracy} (CRA) requires every address in the ordered code to be realized correctly.

Conventional MNIST$\rightarrow$SVHN coherence asks whether an MNIST  image of digit $k$ generates an SVHN image that an independent  classifier recognizes as the same digit $k$. CIIR decomposes this  evaluation into:    
\begin{itemize}
\item \textbf{Address Routing Accuracy:}MNIST image $\rightarrow$ selected address.      It measures whether the source image selects the correct address $k$.        
\item \textbf{Target-side ARA:}      correct address $k$ $\rightarrow$ generated SVHN image $\rightarrow$ independent classifier. It measures whether the requested address is correctly realized as digit $k$.
\item \textbf{End-to-end Address Transfer Accuracy:}MNIST image $\rightarrow$ selected address $\rightarrow$ generated SVHN image $\rightarrow$ independent classifier. It requires both correct address selection and correct image realization.  
\end{itemize}

This decomposition distinguishes address-selection errors from  generated-image realization errors.

CIIR is evaluated here through generative receivers that decode an
address representation into an image or spectrogram. Independent judges
classify these outputs solely for evaluation, so ARA measures successful
address communication rather than direct source classification.
Generation provides an inspectable test that the representation can be
interpreted in each module's native output domain. The interface is not
restricted to generation; receivers could instead return class scores,
retrieval results, control actions, or other address-conditioned outputs.

\subsection{Learning the Address Space}
\label{sec:core_ciir}

We construct the base address space from MNIST, Fashion-MNIST, FSDD, and Sign Language MNIST. For this base configuration, each dataset uses ten local indices, $k\in\{0,\ldots,9\}$, enabling a common one-to-one mapping from local index $k$ to address $k$. The equal index count is a structural choice for learning the shared interface and does not imply that equal indices share semantics across datasets. MNIST, Fashion-MNIST, and Sign Language MNIST use $28\times28\times1$ inputs, whereas FSDD uses $36\times36\times1$ mel-spectrograms.

Each dataset first learns a private encoder--decoder pair. After this stage, the private encoders are discarded and the dataset-specific decoders $D_m$ are frozen. A unified encoder $U$ accepts variable spatial dimensions and produces $K$ address representations and auxiliary logits,
\begin{equation}
U(x) = (Z(x), q(x)),
\qquad
Z(x)\in\mathbb{R}^{K\times d},
\end{equation}
with $K=10$ and $d=128$.

For a sample with local index $k$, CIIR selects the corresponding address representation $Z_k(x)$. Although $U$ produces all $K$ address representations, only $Z_k(x)$ participates in the decoder path for that sample. The unified encoder receives only the observation $x$; the local index is not concatenated with its input and is used only to select the address head. Thus, the index specifies where information is written, whereas the observation determines the continuous representation written there. The auxiliary head $q(x)$ provides a training signal and is not used for address selection in the reported base experiments.

During address-space learning, a sample with local index $k$ writes to address $k$. The frozen source decoder reconstructs the source observation, while two frozen target decoders are sampled from the remaining modalities to encourage cross-module readability at the same address. With $\mathcal{N}_m$ denoting the two target modalities sampled for source modality $m$, the objective is
\begin{equation}
\mathcal{L}_{\mathrm{CIIR}}
=
\mathcal{L}_{\mathrm{self}}
+
\lambda_{\mathrm{idx}}\mathcal{L}_{\mathrm{idx}}
+
\sum_{n\in\mathcal{N}_m}
\mathcal{L}_{\mathrm{cross}}^{(m\rightarrow n)}.
\label{eq:ciir_loss}
\end{equation}
We fix $\lambda_{\mathrm{idx}}=0.5$ across all reported experiments. This downweights the auxiliary index loss so that it regularizes address organization without dominating the self- and cross-module reconstruction objectives.

The original decoders receive the local index because it is part of their training interface. We therefore do not treat this stage alone as evidence that the continuous address representation carries class identity; this question is tested separately with label-free receivers. CIIR assumes that the local index is available during base address-space training and address selection. It therefore operates as a modular communication framework rather than an autonomous classifier or open-set routing method. Larger or differently structured class spaces are handled separately through multi-address codes. Exact architectures, preprocessing, and loss weights are provided in the technical supplement.

\subsection{Base Address-Space Evaluation Protocols}
\label{sec:base_interface_eval}

\paragraph{Label-free receiver evaluation.}
To test whether the learned address representations communicate the requested address independently of decoder-side labels, we freeze $U$ and train one label-free receiver per dataset:
\begin{equation}
\hat{x}^{(m)}
=
D_m^{\mathrm{free}}\!\left(Z_k(x^{(m)})\right).
\label{eq:base_label_free_realization}
\end{equation}
Each receiver receives only the selected 128-dimensional address representation and produces an output in its native domain. An independent judge classifies the output solely for evaluation. We report ARA, SSIM, L1, and a task-specific diversity-retention measure on samples held out from receiver and judge training. We define diversity retention as the ratio of generated within-class pixel variance to the corresponding within-class variance in real test samples. A value of one indicates matched variation, whereas a value near zero indicates that the receiver produces nearly identical outputs within an address. For FSDD, real recordings are split before augmentation. 

To compare input-dependent representations with a fixed reference shared across all receivers, we construct a single address-prototype bank from MNIST:
\begin{equation}
p_k
=
\frac{1}{|\mathcal{B}_k|}
\sum_{i\in\mathcal{B}_k}
Z_k(x_i^{\mathrm{MNIST}}),
\label{eq:prototype_bank}
\end{equation}
where $\mathcal{B}_k$ contains up to 5,000 randomly sampled MNIST training observations with local index $k$. MNIST provides one consistent reference for the ten-address vocabulary, allowing the same prototype bank to be evaluated through every modality-specific receiver and reused unchanged in the later SVHN and CIFAR-100 experiments. Within each modality, $Z_k(x)$ and $p_k$ are passed through the same frozen label-free receiver and evaluated with the same independent judge. For MNIST, this is a same-domain comparison. For the other datasets, it measures how well a receiver realizes a fixed prototype derived from the common reference bank.

\paragraph{Learned fixed address-vector baselines.}
To test whether CIIR can be replaced by one learned vector per address, we compare its input-dependent address representations with two lookup baselines on MNIST. Both use $E\in\mathbb{R}^{10\times128}$ and return $E_k$ for the supplied local index. In FV-FR (fixed address vector, 'frozen' label-free receiver), only the lookup table is trained while the label-free receiver remains frozen. In FV-TR (fixed address vector, 'fine-tuned' label-free receiver), the lookup table and a receiver initialized from the saved CIIR receiver are fine-tuned jointly. Both baselines use the MNIST training split and are evaluated on the official test split with the same judge and metrics. We report mean and standard deviation over three random seeds.

\subsection{Post-hoc Attachment of SVHN}
\label{sec:svhn_attachment}

After base-address-space learning, all original CIIR modules are frozen. We reuse the MNIST-derived address prototypes in
Eq.~\ref{eq:prototype_bank} as
\begin{equation}
P=\{p_0,\ldots,p_9\}.
\label{eq:address_bank}
\end{equation}
A label-free SVHN autoencoder defines a private 128-dimensional SVHN latent space. A source-to-address routing head predicts address $\hat{k}$ from the frozen MNIST encoder outputs. A write adapter maps an address prototype and noise to the SVHN latent space,
\begin{equation}
\hat z_{\mathrm{SVHN}}=W(p_k,\epsilon),
\end{equation}
and a read adapter predicts an address from an SVHN latent vector,
\begin{equation}
\hat{k}=\arg\max R(z_{\mathrm{SVHN}}).
\end{equation}
The forward and reverse paths are
\begin{equation}
\mathrm{MNIST}\rightarrow p_{\hat{k}}\rightarrow W
\rightarrow D_{\mathrm{SVHN}}^{\mathrm{free}},
\end{equation}
and
\begin{equation}
\mathrm{SVHN}\rightarrow R\rightarrow p_{\hat{k}}
\rightarrow D_{\mathrm{MNIST}}^{\mathrm{free}}.
\end{equation}
Neither target decoder receives a class label. We report Address Routing Accuracy, target-side ARA, End-to-end Address Transfer Accuracy, Transfer Retention, and address-representation interventions. Adapter losses and SVHN training details are given in the supplement.

\paragraph{Attaching additional datasets.}
To attach a new dataset $r$, CIIR freezes the existing system and trains a native label-free output module together with a write adapter $W_r$ and read adapter $R_r$. The current design therefore requires one dataset-specific adapter pair per added dataset, but no retraining of the base address space or previous adapters.

\subsection{CIFAR-100 Address-Code Evaluation}
\label{sec:cifar_composition}

CIFAR-100 is used to test reuse of the ten-address vocabulary for a larger class space, rather than CIFAR-100 image generation, classification, or semantic alignment. Each of its 100 known fine-class identities is represented as an ordered sequence of existing CIIR addresses. For fine class $c$,
\begin{equation}
\phi_{\mathrm{arith}}(c)
=
\left[
\left\lfloor \frac{c}{10}\right\rfloor,
c\bmod 10
\right].
\label{eq:arith_code}
\end{equation}
For superclass $s(c)\in\{0,\ldots,19\}$ and within-superclass member
$m(c)\in\{0,\ldots,4\}$,
\begin{equation}
\phi_{\mathrm{hier}}(c)
=
\left[
\left\lfloor\frac{s(c)}{10}\right\rfloor,
s(c)\bmod 10,
m(c)
\right].
\label{eq:hier_code}
\end{equation}

We evaluate the resulting codes at two levels. First, we evaluate their deterministic structure independently of any receiver. Round-trip Recovery Accuracy measures whether decoding an address code returns its original CIFAR-100 fine-class identity. Collision Count records the number of fine-class pairs assigned the same code. For hierarchical codes, Superclass Prefix Consistency measures whether all fine classes within a superclass share the same first two addresses.

Second, we evaluate whether the addresses in a code can be realized through the existing modality-specific receivers. For a code $\phi(c)=[k_1,\ldots,k_L]$, each requested address $k_j$ retrieves the frozen MNIST-derived prototype $p_{k_j}$, which is passed to a label-free receiver. An independent judge then predicts the realized address $\hat{k}_j$. ARA measures the fraction of individual code positions realized correctly, whereas CRA measures the fraction of CIFAR-100 classes for which the complete ordered code is realized correctly. This receiver-level evaluation executes a known class code through the existing CIIR interface; it does not classify or reconstruct the CIFAR-100 image.

\section{Results}
\label{sec:results}

\subsection{Base Address-Space Evaluation}
\label{sec:results_base_interface}

\paragraph{Label-free receiver evaluation.}
Table~\ref{tab:label_free_representation} shows address realization through receivers that receive no address label. Input-dependent address representations achieved 94.7\% macro-average ARA across the four datasets, ranging from 89.1\% on FSDD to 98.6\% on MNIST. Compared with the fixed MNIST-derived address prototypes, they also preserved substantially greater reconstruction fidelity and within-class diversity. Div. Ret. 0.00 indicates that a fixed prototype generates essentially the
same output for every sample assigned to that address. Qualitative outputs from the original label-conditioned decoders are shown in Fig.~\ref{fig:base10_all_modalities}.

Address-representation substitution further verified control by the continuous representation. Replacing the original representation with one from another address redirected the output to the substituted address with 94.7\% macro-average ARA, while the original address was retained in only 0.2\% of cases. Zero and random representations remained near chance at 10.0\% and 10.3\%. Full results are provided in the supplement.

\begin{table}[t]
\centering
\fontsize{7}{8}\selectfont
\setlength{\tabcolsep}{1.8pt}
\renewcommand{\arraystretch}{1.05}
\caption{ 
Label-free receiver evaluation. Input-dependent address representations are compared with a fixed MNIST-derived address-prototype bank. Address Realization Accuracy (ARA) measures realization of the requested address. Div. Ret. denotes diversity retention (generated within-class variance relative to real test data; 0.0000 indicates that a fixed prototype generates essentially the same
output for every sample assigned to that address).
}
\label{tab:label_free_representation}
\begin{tabular}{@{}llcccc@{}}
\toprule
\textbf{Dataset}
& \textbf{Representation}
& \textbf{ARA}
& \textbf{SSIM}
& \textbf{L1}
& \textbf{Div. Ret.} \\
\midrule
MNIST
& Input-dependent $Z_k(x)$
& 0.9855 & 0.8434 & 0.0435 & 0.9055 \\
& MNIST prototype $p_k$
& 1.0000 & 0.6511 & 0.1097 & 0.0000 \\
\addlinespace
Fashion-MNIST
& Input-dependent $Z_k(x)$
& 0.9343 & 0.6943 & 0.0597 & 0.8609 \\
& MNIST prototype $p_k$
& 0.8000 & 0.3498 & 0.2238 & 0.0000 \\
\addlinespace
FSDD
& Input-dependent $Z_k(x)$
& 0.8911 & 0.7518 & 0.0420 & 0.8552 \\
& MNIST prototype $p_k$
& 0.4000 & 0.2066 & 0.2192 & 0.0000 \\
\addlinespace
Sign Language MNIST
& Input-dependent $Z_k(x)$
& 0.9762 & 0.9263 & 0.0231 & 0.9266 \\
& MNIST prototype $p_k$
& 0.1000 & 0.1708 & 0.2101 & 0.0000 \\
\midrule
\textbf{Macro average}
& Input-dependent $Z_k(x)$
& \textbf{0.9468} & \textbf{0.8039}
& \textbf{0.0421} & \textbf{0.8870} \\
& MNIST prototype $p_k$
& 0.5750 & 0.3446 & 0.1907 & 0.0000 \\
\bottomrule
\end{tabular}
\end{table}

\paragraph{Original label-conditioned decoder evaluation.}
Table~\ref{tab:conditioned_representation} shows reconstruction through the original label-conditioned decoders using either an input-dependent address representation or a modality-specific address prototype. Because the known address $k$ is supplied in both conditions, ARA mainly
measures output recognizability. Input-dependent representations increased macro-average SSIM from 0.4927 to 0.5278, reduced L1 error from 0.1341 to 0.1132, and retained 19.2\% of real within-class
variation, whereas the fixed prototypes retained none. In other words, the prototype produces nearly the same class-representative output for all samples, while $Z_k(x)$ preserves sample-specific variation.

\begin{table}[t]
\centering
\fontsize{7}{8}\selectfont
\setlength{\tabcolsep}{1.8pt}
\renewcommand{\arraystretch}{1.05}
\caption{
Original label-conditioned decoder evaluation. Input-dependent address representations are compared with modality-specific address prototypes. Because the known address $k$ is supplied in both conditions, SSIM, L1, and diversity retention provide the primary evidence of observation-specific information. Div. Ret. denotes diversity retention relative to real test data; 0.0000 indicates that a fixed prototype produces essentially the same output for every sample assigned to that address.
}
\label{tab:conditioned_representation}
\begin{tabular}{@{}llcccc@{}}
\toprule
\textbf{Dataset}
& \textbf{Representation}
& \textbf{ARA}
& \textbf{SSIM}
& \textbf{L1}
& \textbf{Div. Ret.} \\
\midrule
MNIST
& Input-dependent $Z_k(x)$
& 0.7730 & \textbf{0.6566} & \textbf{0.1051}
& \textbf{0.1550} \\
& Modality prototype $p_k^{(m)}$
& \textbf{0.9018} & 0.6474 & 0.1097 & 0.0000 \\
\addlinespace
Fashion-MNIST
& Input-dependent $Z_k(x)$
& 0.9752 & \textbf{0.5283} & \textbf{0.1067}
& \textbf{0.3412} \\
& Modality prototype $p_k^{(m)}$
& \textbf{1.0000} & 0.4755 & 0.1451 & 0.0000 \\
\addlinespace
FSDD
& Input-dependent $Z_k(x)$
& \textbf{0.8444} & \textbf{0.4265} & \textbf{0.1036}
& \textbf{0.2117} \\
& Modality prototype $p_k^{(m)}$
& 0.8000 & 0.3549 & 0.1330 & 0.0000 \\
\addlinespace
Sign Language MNIST
& Input-dependent $Z_k(x)$
& 0.3734 & \textbf{0.4997} & \textbf{0.1375}
& \textbf{0.0590} \\
& Modality prototype $p_k^{(m)}$
& \textbf{0.4000} & 0.4931 & 0.1488 & 0.0000 \\
\midrule
\textbf{Macro average}
& Input-dependent $Z_k(x)$
& 0.7415 & \textbf{0.5278} & \textbf{0.1132}
& \textbf{0.1917} \\
& Modality prototype $p_k^{(m)}$
& \textbf{0.7755} & 0.4927 & 0.1341 & 0.0000 \\
\bottomrule
\end{tabular}
\end{table}

\begin{table}[t]
\centering
\fontsize{7}{8}\selectfont
\setlength{\tabcolsep}{4.5pt}
\renewcommand{\arraystretch}{1.10}
\caption{
Learned fixed address-vector baselines on MNIST. CIIR uses an input-dependent address representation, whereas FV-FR and FV-TR use one fixed learned vector per address. CIIR reports the saved-model result; the baselines report mean $\pm$ standard deviation over three seeds. 
}
\label{tab:fixed_address_vectors}
\begin{tabular}{@{}lcccc@{}}
\toprule
\textbf{Rep.}
& \textbf{ARA}
& \textbf{SSIM}
& \textbf{L1}
& \textbf{Div. Ret.} \\
\midrule
CIIR\textsuperscript{1}
& \textbf{0.9855}
& \textbf{0.8434}
& \textbf{0.0435}
& \textbf{0.9055} \\
FV-FR\textsuperscript{2}
& $0.7698{\pm}0.1104$
& $0.6771{\pm}0.0013$
& $0.1052{\pm}0.0012$
& $\approx 0$ \\
FV-TR\textsuperscript{3}
& $0.8397{\pm}0.0616$
& $0.6759{\pm}0.0044$
& $0.1069{\pm}0.0028$
& $\approx 0$ \\
\bottomrule
\end{tabular}
\vspace{1mm}
\begin{minipage}{\columnwidth}
\fontsize{7}{8}\selectfont
\textsuperscript{1}CIIR: input-dependent address representation
$Z_k(x)$ produced from each observation.
\textsuperscript{2}FV-FR: learned fixed address vector with the
label-free receiver frozen.
\textsuperscript{3}FV-TR: learned fixed address vector with the
label-free receiver fine-tuned jointly.
Div. Ret.: diversity retention relative to real test data.
\end{minipage}
\end{table}

\begin{table*}[!t]
\centering
\fontsize{7}{8}\selectfont
\setlength{\tabcolsep}{3pt}
\renewcommand{\arraystretch}{1.06}
\caption{
Reported results on the MNIST--SVHN--Text multimodal representation-learning benchmark \cite{sutter2020mmjsd}. Prior methods report their original evaluation metrics, while CIIR reports End-to-end Address Transfer Accuracy and target-side ARA through the source$\rightarrow$address$\rightarrow$target path.
}
\label{tab:mnist_svhn_all}
\begin{tabular*}{\textwidth}{
@{\extracolsep{\fill}}
p{0.16\textwidth}
p{0.23\textwidth}
p{0.17\textwidth}
p{0.20\textwidth}
p{0.14\textwidth}
@{}
}
\toprule
\textbf{Model}
& \textbf{Intermediate Representation}
& \textbf{Direction}
& \textbf{Metric}
& \textbf{Score} \\
\midrule
MVAE\textsuperscript{1}
& PoE latent
& MNIST $\rightarrow$ SVHN
& Digit matching
& 9.5\% \\
& PoE latent
& SVHN $\rightarrow$ MNIST
& Digit matching
& 9.3\% \\
\midrule
MMVAE\textsuperscript{2}
& MoE shared latent
& MNIST $\rightarrow$ SVHN
& Digit matching
& 86.4\% \\
& MoE shared latent
& SVHN $\rightarrow$ MNIST
& Digit matching
& 69.1\% \\
\midrule
mmJSD\textsuperscript{3}
& JSD-aligned latent
& MST $\rightarrow$ missing modalities
& Conditional coherence
& 0.72 \\
\midrule
DMVAE\textsuperscript{4}
& Shared/private latent
& MNIST $\rightarrow$ SVHN
& Cross-synthesis accuracy
& 88.1\% \\
& Shared/private latent
& SVHN $\rightarrow$ MNIST
& Cross-synthesis accuracy
& 83.7\% \\
\midrule
MoPoE-VAE\textsuperscript{5}
& Mixture-of-products latent
& MST $\rightarrow$ missing modalities
& Conditional coherence
& 0.69 \\
& Mixture-of-products latent
& Prior $\rightarrow$ MST modalities
& Unconditional coherence
& 0.31 \\
\midrule
JNF-Shared (CL)\textsuperscript{6}
& Shared latent + flow posterior
& MNIST $\rightarrow$ SVHN
& Conditional coherence
& 0.81 \\
& Shared latent + flow posterior
& SVHN $\rightarrow$ MNIST
& Conditional coherence
& 0.75 \\
& Shared latent + flow posterior
& Prior $\rightarrow$ MNIST/SVHN
& Joint coherence / FID
& 0.51 / 49 \\
\midrule
CoDE-VAE\textsuperscript{7}
& Dependent-expert latent
& MST $\rightarrow$ missing modalities
& Conditional coherence
& 0.82 \\
& Dependent-expert latent
& MNIST/SVHN/Text $\rightarrow$ class
& Latent classification
& 0.95 \\
& Dependent-expert latent
& Conditional input $\rightarrow$ target
& Conditional FID
& 83.6\textsuperscript{$\dagger$} \\
\midrule
ShaLa\textsuperscript{8}
& Shared latent + diffusion prior
& MST $\rightarrow$ missing modalities
& Conditional coherence
& 0.75 \\
& Shared latent + diffusion prior
& Prior $\rightarrow$ MST modalities
& Unconditional coherence
& 0.44 \\
\midrule
\multirow[t]{4}{*}{Iterative Amortized Inference\textsuperscript{9}}
& Refined unimodal latent
& MNIST $\rightarrow$ SVHN
& Cross-coherence
& 82.11\% \\
& Refined unimodal latent
& SVHN $\rightarrow$ MNIST
& Cross-coherence
& 75.63\% \\
& Refined unimodal latent
& MNIST $\rightarrow$ SVHN
& FID
& 49.43 \\
& Refined unimodal latent
& SVHN $\rightarrow$ MNIST
& FID
& 50.13 \\
\midrule
H\"older++\textsuperscript{10}
& Shared latent
& MNIST $\rightarrow$ class
& Latent classification
& 97.0\% \\
& Shared latent
& SVHN $\rightarrow$ class
& Latent classification
& 92.2\% \\
& Joint latent
& MNIST/SVHN $\rightarrow$ class
& Joint-latent classification
& 97.7\% / 92.2\% \\
\midrule
CIIR (Ours)
&  CIIR address space
&  MNIST $\rightarrow$ SVHN
&  End-to-end Address Transfer Accuracy
&  90.24\% \\
&  CIIR address space
&  SVHN $\rightarrow$ MNIST
&  End-to-end Address Transfer Accuracy
&  82.29\% \\
&  CIIR address space
&  Correct address $\rightarrow$ SVHN
&  Target ARA
&  99.18\% \\
&  CIIR address space
&  Correct address $\rightarrow$ MNIST
&  Target ARA
&  100.0\% \\
&  CIIR address space
&  MNIST $\rightarrow$ SVHN
&  FID
&  62.50 \\
\bottomrule
\end{tabular*}

\vspace{1mm}
\begin{minipage}{\textwidth}
\fontsize{7}{8}\selectfont
\textsuperscript{1}\cite{wu2018mvae};
\textsuperscript{2}\cite{shi2019mmvae};
\textsuperscript{3}\cite{sutter2020mmjsd};
\textsuperscript{4}\cite{lee2021dmvae};
\textsuperscript{5}\cite{sutter2021mopoe};
\textsuperscript{6}\cite{senellart2025inferencegap};
\textsuperscript{7}\cite{mancisidor2025code};
\textsuperscript{8}\cite{cui2026shala};
\textsuperscript{9}\cite{oshima2026enhancing};
\textsuperscript{10}\cite{vo2026holderpp}.

\textsuperscript{$\dagger$}The CoDE-VAE FID corresponds to a reported parameter-sweep configuration. JNF-Shared (CL) is one of several variants reported in the cited work. Results should be interpreted under each method's original evaluation protocol.
\end{minipage}
\end{table*}

\paragraph{Learned fixed address-vector baselines.}
Table~\ref{tab:fixed_address_vectors} compares CIIR with learned fixed address vectors on MNIST. FV-FR (learned fixed address vector with the label-free receiver frozen) and FV-TR (learned fixed address vector with the label-free receiver fine-tuned jointly) reached 77.0\% and 84.0\% mean ARA, respectively, compared with 98.6\% for CIIR. CIIR also achieved substantially higher SSIM and lower L1 error, indicating more faithful reconstruction. Most importantly, both fixed-vector baselines retained effectively no within-class variation, even when the receiver was fine-tuned jointly. Thus, a learned vector can encode a recognizable class prototype, but it cannot reproduce the sample-specific information carried by the input-dependent representation $Z_k(x)$.

\subsection{Post-hoc SVHN Attachment}
\label{sec:results_svhn}

Table~\ref{tab:mnist_svhn_all} evaluates whether a source observation retains its class-indexed address through the complete source$\rightarrow$address$\rightarrow$target path. For example, an MNIST image with digit index $0$ must be assigned to address $0$, mapped through the write adapter, and decoded as an SVHN image that an independent judge recognizes as digit $0$. End-to-end Address Transfer Accuracy counts a sample as correct only when this address is preserved through the complete path.

For MNIST$\rightarrow$SVHN, the source image is first assigned to an address by the routing head. The selected address prototype is then mapped through the write adapter and decoded by the label-free SVHN receiver. Address Routing Accuracy was 90.95\%, while target-side ARA was 99.18\% when the correct address was supplied. The resulting End-to-end Address Transfer Accuracy was 90.24\%.
The small difference between routing accuracy and end-to-end accuracy shows that nearly all observed error arose during source-side address assignment rather than target-side realization.

For SVHN$\rightarrow$MNIST, the read adapter assigned the correct address in 82.29\% of cases. The label-free MNIST receiver achieved 100\% target-side ARA for correctly selected addresses, yielding the same 82.29\% End-to-end Address Transfer Accuracy. In this direction, all observed end-to-end errors therefore occurred during SVHN-to-address assignment rather than target realization.

\begin{figure}[!t]
\centering
\fontsize{7}{8}\selectfont
\includegraphics[width=\columnwidth]
{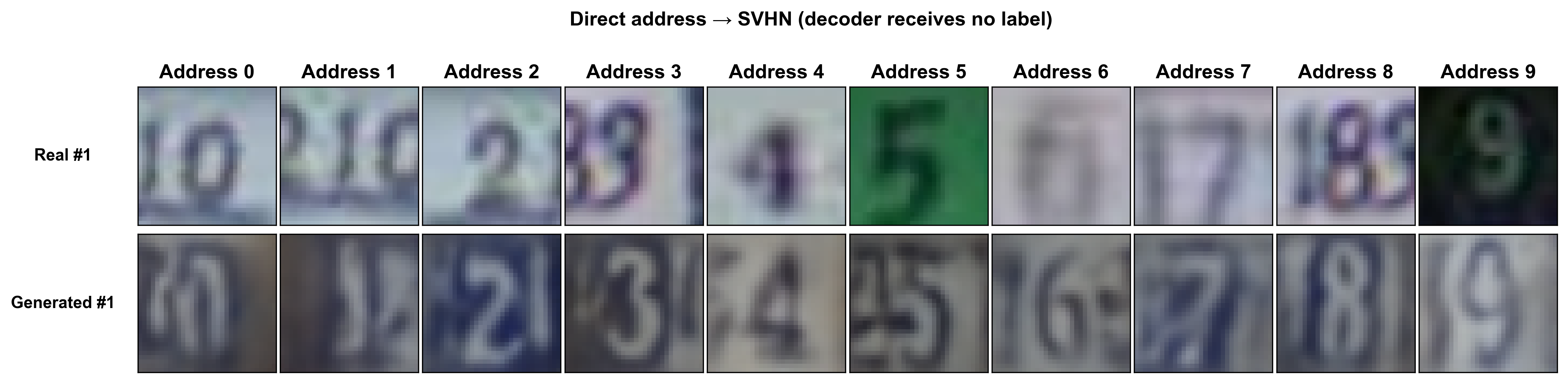}
\caption{
Real SVHN examples and label-free outputs generated from the ten frozen
CIIR address prototypes. The SVHN decoder receives no target-side class
label.
}
\label{fig:svhn_address_realization}
\end{figure}

The closest direction-specific results from prior multimodal methods are included in Table~\ref{tab:mnist_svhn_all}. CIIR reaches comparable or higher address-preservation rates while attaching SVHN after the base address space has been frozen and training only the new routing and read/write components. Because prior methods report related but not identical generation-consistency metrics, the table provides contextual rather than strictly metric-identical comparison. Qualitative SVHN realizations are shown in Fig.~\ref{fig:svhn_address_realization}.

\begin{table}[!t]
\centering
\fontsize{7}{8}\selectfont
\setlength{\tabcolsep}{9.5pt}
\renewcommand{\arraystretch}{1.06}
\caption{
CIFAR-100 address-code evaluation. RT and Collision describe the deterministic code structure. ARA measures realization of individual addresses, and CRA requires every address in a class code to be realized correctly.
}
\label{tab:cifar_code_execution}
\begin{tabular}{@{}llcccc@{}}
\toprule
\textbf{Code}
& \textbf{Receiver}
& \textbf{RT}
& \textbf{Collision}
& \textbf{ARA}
& \textbf{CRA} \\
\midrule
Arithmetic
& MNIST
& 1.000 & 0 & 1.000 & 1.000 \\
& F-MNIST
& 1.000 & 0 & 0.800 & 0.640 \\
& FSDD
& 1.000 & 0 & 0.400 & 0.160 \\
& Sign-MNIST
& 1.000 & 0 & 0.100 & 0.010 \\
\addlinespace
Hierarchical
& Structure
& 1.000 & 0 & -- & -- \\
\bottomrule
\end{tabular}

\vspace{1mm}
\begin{minipage}{\columnwidth}
\fontsize{7}{8}\selectfont
RT: Round-trip Recovery Accuracy. Collision: Collision Count. ARA: Address Realization Accuracy. CRA: Class-Code Realization Accuracy. F-MNIST: Fashion-MNIST. Sign-MNIST: Sign Language MNIST. Structure: deterministic structural analysis without receiver execution.
\end{minipage}
\end{table}

\subsection{CIFAR-100 Address-Code Evaluation}
\label{sec:results_cifar100}
Table~\ref{tab:cifar_code_execution} evaluates the CIFAR-100$\rightarrow$address-code$\rightarrow$receiver pipeline. Each known CIFAR-100 class is first mapped to an ordered code over the ten CIIR addresses, and the corresponding MNIST-derived address prototypes are then realized through the four modality-specific receivers. Both coding schemes achieved 100\% Round-trip Recovery Accuracy with zero collisions, and the hierarchical
scheme achieved 100\% Superclass Prefix Consistency. For arithmetic codes, ARA/CRA were 100.0/100.0\% for MNIST, 80.0/64.0\% for Fashion-MNIST, 40.0/16.0\% for FSDD, and 10.0/1.0\% for Sign Language MNIST. The lower non-MNIST scores reflect receiver compatibility with the MNIST-derived prototype bank, not errors in the 100-class code assignment. Example arithmetic and hierarchical code realizations are shown in Fig.~\ref{fig:cifar_code_realizations}.

\section{Discussion}

CIIR connects independently collected datasets through a shared,
class-indexed address space while retaining their modality-specific
decoders. Label-free receivers, representation substitutions, original
decoder evaluation, and fixed-vector baselines show that the learned
address representations carry both class-readable and input-specific
information beyond fixed address vectors. Post-hoc SVHN attachment
further shows that a frozen address space can incorporate a new dataset
through lightweight read/write adapters, achieving 90.24\%
MNIST$\rightarrow$SVHN and 82.29\% SVHN$\rightarrow$MNIST
End-to-end Address Transfer Accuracy, with target-side ARA of 99.18\%
and 100\%, respectively. For CIFAR-100, both arithmetic and hierarchical
codes achieved exact round-trip recovery with zero collisions, although
receiver-level realization depended on compatibility with the
MNIST-derived prototype bank.

The current framework assumes predefined mappings from dataset-local
class indices to the shared address vocabulary. Larger or differently
structured class spaces therefore still require manually specified
multi-address codes. Future work should learn hybrid codes that combine
class-index addresses with reusable semantic components while preserving
CIIR's modular interface.

\section{Conclusion}

CIIR learns input-dependent representations in a shared class-indexed
space for communication among independently collected datasets.
Label-free realization, representation substitutions, post-hoc SVHN
attachment, and CIFAR-100 address coding show that the interface supports
modular generation and reuse without fixed address vectors or exhaustive
paired multimodal supervision.

CIIR is particularly relevant when deployed modules must remain frozen
because they have been validated, are maintained independently, or were
trained on data that cannot be pooled. In industrial, medical, robotic,
and multi-sensor systems, existing part, diagnosis, fault, or task
identifiers can serve as shared addresses. Similar to trigger--action
integration in mobile and IoT platforms, CIIR allows one module's
recognized state to activate generation, retrieval, diagnosis, or control
in another through lightweight interfaces rather than full-system
retraining.

\bibliography{aaai2027}
\clearpage 
\newpage

\section{Technical Supplement}
\label{sup:technical}

\subsection{Network Architectures}
\label{sup:architectures}

All models are implemented in TensorFlow/Keras. CIIR does not use a
standard ResNet backbone. Instead, it uses convolutional encoders whose
spatial outputs are reduced by global average pooling. Consequently, the
same unified encoder accepts both $28\times28$ images and $36\times36$
spectrograms without spatial resizing inside the model. All convolutional
layers use same padding unless stated otherwise.

\paragraph{Private encoders.}
Each base dataset first trains a private encoder with three
$3\times3$ stride-2 convolutional layers containing 64, 128, and 256
channels. Every convolution is followed by LeakyReLU with negative slope
0.2 and batch normalization. Global average pooling is followed by a
256-unit ReLU layer and a linear 128-dimensional latent layer. The
private encoders are discarded after training.

\paragraph{Dataset-specific decoders.}
Each original decoder receives a 128-dimensional private latent vector and a
64-dimensional embedding of the known local index. Their concatenation
is projected to a spatial tensor of size
$(H/4)\times(W/4)\times256$, where $(H,W)$ is the native output
resolution. Two $3\times3$ stride-2 transposed convolutions with 256 and
128 channels restore the native resolution. These are followed by
$3\times3$ convolutions with 128, 64, and 32 channels and a sigmoid
output layer. Batch normalization and LeakyReLU are used throughout.
MNIST, Fashion-MNIST, and Sign Language MNIST use $28\times28\times1$
outputs; FSDD uses $36\times36\times1$ outputs.

\paragraph{Unified encoder.}
The unified encoder receives an input of shape
$H\times W\times1$, where $H$ and $W$ may vary across datasets. Its
convolutional stack is
\[
64
\rightarrow
64_{\downarrow2}
\rightarrow
128
\rightarrow
128_{\downarrow2}
\rightarrow
256_{\downarrow2},
\]
using $3\times3$ kernels. Each layer is followed by LeakyReLU with slope
0.2 and batch normalization. Global average pooling is followed by
512- and 256-unit ReLU layers. One linear head produces ten auxiliary
index logits. Ten separate address heads each apply a 256-unit ReLU layer
followed by a linear 128-dimensional output, producing the address-representation tensor
$Z(x)\in\mathbb{R}^{10\times128}$.

\paragraph{Label-free MNIST receiver.}
The MNIST receiver takes only a 128-dimensional address representation. A fully connected
layer produces a $7\times7\times256$ tensor, followed by two stride-2
transposed convolutions with 256 and 128 channels. A 64-channel
convolution and a one-channel sigmoid convolution produce the
$28\times28$ output. No class label is supplied.

\paragraph{SVHN encoder and receiver.}
The SVHN encoder uses three $3\times3$ stride-2 convolutions with 64,
128, and 256 channels, each followed by batch normalization and
LeakyReLU. Global average pooling is followed by 512- and 256-unit ReLU
layers. As in the base unified encoder, it has ten 128-dimensional address heads and a ten-way auxiliary classification head.

The label-free SVHN receiver maps a 128-dimensional address representation to a
$4\times4\times256$ tensor and applies three $4\times4$ stride-2
transposed convolutions with 256, 128, and 64 channels. A 64-channel
convolution and a three-channel sigmoid output produce a
$32\times32\times3$ image. No target-side class label is provided.

\paragraph{SVHN attachment modules.}
The MNIST routing head flattens the ten 128-dimensional unified-encoder address representations, concatenates them with the ten auxiliary logits, and applies
512- and 256-unit ReLU layers with dropout 0.25 before a ten-way output.

The SVHN write adapter concatenates a 128-dimensional address prototype with 128-dimensional Gaussian noise. It applies ReLU layers
of size 512, 512, and 256, followed by a linear 128-dimensional SVHN
latent output. The read adapter maps a selected 128-dimensional SVHN latent vector through 256- and 128-unit ReLU layers with dropout 0.2 and produces
ten address logits.

\paragraph{Exploratory CIFAR-100 expansion encoder.}
The exploratory expansion encoder receives a grayscale
$28\times28\times1$ CIFAR-100 image and a 64-dimensional embedding of
its known fine-class identity. Its convolutional image branch follows
the same channel sequence as the unified encoder:
$64$, $64_{\downarrow2}$, $128$, $128_{\downarrow2}$, and
$256_{\downarrow2}$, with batch normalization and LeakyReLU. Global
average pooled image features are concatenated with the class embedding,
then passed through 512- and 256-unit ReLU layers. Two independent linear heads produce candidate address representations
$\widetilde{Z}_{k_1}(x,c)$ and $\widetilde{Z}_{k_2}(x,c)$ for the two
requested addresses in the arithmetic code. This encoder is evaluated only in the supplementary exploratory analysis
and is not used for the direct address-prototype-bank results reported in
Table~\ref{tab:cifar_code_execution}.

\begin{table*}[t]
\centering
\fontsize{7}{8}\selectfont
\caption{
Architectures of the principal CIIR components. All convolutional
kernels are $3\times3$ unless otherwise specified. GAP denotes global
average pooling, BN denotes batch normalization, and LReLU uses slope
0.2.
}
\label{sup:tab:architectures}
\setlength{\tabcolsep}{3pt}
\renewcommand{\arraystretch}{1.08}
\begin{tabular}{@{}p{0.16\textwidth}p{0.64\textwidth}p{0.12\textwidth}@{}}
\toprule
\textbf{Module} & \textbf{Architecture} & \textbf{Output} \\
\midrule
Private encoder
&
Conv-64/2--LReLU--BN;
Conv-128/2--LReLU--BN;
Conv-256/2--LReLU--BN;
GAP; FC-256-ReLU
&
128-D
\\

Original decoder
&
$[h^{(m)};\mathrm{Emb}_{64}(k)]$;
FC-$(H/4)(W/4)256$;
Deconv-256/2;
Deconv-128/2;
Conv-128--Conv-64--Conv-32
&
$H\times W\times C$
\\

Unified encoder
&
Conv-64;
Conv-64/2;
Conv-128;
Conv-128/2;
Conv-256/2;
GAP; FC-512; FC-256;
10 independent FC-256 address heads
&
$10\times128$ representations
+ 10 logits
\\

MNIST label-free receiver
&
FC-$7{\times}7{\times}256$;
Deconv-256/2;
Deconv-128/2;
Conv-64
&
$28\times28\times1$
\\

SVHN encoder
&
Conv-64/2;
Conv-128/2;
Conv-256/2;
GAP; FC-512; FC-256;
10 independent FC-256 address heads
&
$10\times128$ representations
+ 10 logits
\\

SVHN label-free receiver
&
FC-$4{\times}4{\times}256$;
Deconv-256/2;
Deconv-128/2;
Deconv-64/2;
Conv-64
&
$32\times32\times3$
\\

Write adapter
&
$[p_k;\epsilon]$;
FC-512--FC-512--FC-256
&
128-D SVHN latent
\\

Read adapter
&
FC-256--Dropout(0.2)--FC-128
&
10 logits
\\

CIFAR expansion encoder
&
Unified-style convolutional branch;
64-D class embedding;
FC-512--FC-256;
two linear heads
&
$\widetilde{Z}_{k_1},\widetilde{Z}_{k_2}\in\mathbb{R}^{128}$
\\
\bottomrule
\end{tabular}
\end{table*}

\subsection{Training Hyperparameters}
\label{sup:hyperparameters}

We use Adam for all trained components. Unless stated otherwise, batch
normalization parameters are learned jointly with the corresponding
network, and early stopping restores the best observed weights. The
base address space contains $K=10$ addresses with representation dimension $d=128$.
Each base dataset contributes 500 samples per local index, and the base
batch size is 64.

The private autoencoder budgets are dataset specific. MNIST and Sign
Language MNIST are trained for at most 100 epochs with learning rate
$2\times10^{-4}$, a minimum of 40 epochs, and early-stopping patience
20. Fashion-MNIST uses at most 120 epochs with learning rate
$10^{-4}$, minimum 40 epochs, and patience 20. FSDD uses at most 180
epochs with learning rate $10^{-4}$, minimum 60 epochs, and patience
30.

The unified encoder is trained for at most 200 epochs using learning
rate $10^{-4}$, minimum 80 epochs, and patience 25. Every update
processes all four source modalities. For each source modality, two of
the remaining three target decoders are sampled without replacement.
Model selection uses mean cross-module SSIM.

The exploratory CIFAR-100 expansion encoder is trained for at most 80
epochs using Adam with learning rate $2\times10^{-4}$, batch size 16,
minimum 30 epochs, and patience 15. Twenty target examples per local
index and modality are cached for target sampling.

For SVHN attachment, all base CIIR components remain frozen. The MNIST
routing head uses Adam with learning rate $3\times10^{-4}$, batch size
256, at most 35 epochs, and patience 5. The label-free MNIST receiver
uses learning rate $2\times10^{-4}$, batch size 256, at most 60 epochs,
and patience 8. The SVHN encoder and label-free receiver are jointly
trained with Adam at $2\times10^{-4}$ for at most 120 epochs and batch
size 256; training stops after epoch 60 when no improvement is observed
for 14 epochs.

The write adapter uses Adam at $3\times10^{-4}$, batch size 256, and at
most 100 epochs. Each epoch contains 240 randomly sampled mini-batches.
The read adapter uses Adam at $5\times10^{-4}$, batch size 256, at most
40 epochs, and patience 5. Independent judges are trained for at most
30 epochs with batch size 128 and patience 5.

\begin{table*}[t]
\centering
\fontsize{7}{8}\selectfont
\caption{Training and model hyperparameters used in the reported runs.}
\label{sup:tab:hyperparameters}
\setlength{\tabcolsep}{2.5pt}
\renewcommand{\arraystretch}{1.08}
\begin{tabular}{@{}
p{0.20\textwidth}
p{0.10\textwidth}
p{0.13\textwidth}
p{0.07\textwidth}
p{0.13\textwidth}
p{0.24\textwidth}
@{}}
\toprule
\textbf{Stage}
& \textbf{Optimizer}
& \textbf{Learning rate}
& \textbf{Batch}
& \textbf{Maximum epochs}
& \textbf{Early stopping} \\
\midrule
MNIST private AE
& Adam & $2{\times}10^{-4}$ & 64 & 100
& min 40, patience 20 \\
Fashion private AE
& Adam & $1{\times}10^{-4}$ & 64 & 120
& min 40, patience 20 \\
FSDD private AE
& Adam & $1{\times}10^{-4}$ & 64 & 180
& min 60, patience 30 \\
Sign private AE
& Adam & $2{\times}10^{-4}$ & 64 & 100
& min 40, patience 20 \\
Unified encoder
& Adam & $1{\times}10^{-4}$ & 64 & 200
& min 80, patience 25 \\
Exploratory CIFAR encoder
& Adam & $2{\times}10^{-4}$ & 16 & 80
& min 30, patience 15 \\
MNIST router
& Adam & $3{\times}10^{-4}$ & 256 & 35
& patience 5 \\
MNIST label-free receiver
& Adam & $2{\times}10^{-4}$ & 256 & 60
& patience 8 \\
SVHN encoder/receiver
& Adam & $2{\times}10^{-4}$ & 256 & 120
& after 60, patience 14 \\
SVHN write adapter
& Adam & $3{\times}10^{-4}$ & 256 & 100
& after 45, patience 15 \\
SVHN read adapter
& Adam & $5{\times}10^{-4}$ & 256 & 40
& patience 5 \\
Independent judges
& Adam & $5{\times}10^{-4}$ or $10^{-3}$ & 128 & 30
& patience 5 \\
\bottomrule
\end{tabular}
\end{table*}

All reported SVHN attachment runs use random seed 42. Base-address-space and
CIFAR evaluation seeds are reported with their corresponding protocols.
Adam uses the TensorFlow/Keras default momentum parameters unless stated
otherwise, and no explicit weight decay is applied. Dataset split
procedures are described in the evaluation-protocol subsections below.

\subsection{Base Datasets and Preprocessing}
\label{sup:base_data}

The base address space uses MNIST, Fashion-MNIST, FSDD, and Sign Language MNIST.
Each dataset contributes ten local class indices and 500 samples per local class during address-space learning.
Equal index values do not imply shared semantics across datasets.

MNIST, Fashion-MNIST, and Sign Language MNIST use $28\times28\times1$ inputs.
FSDD waveforms are resampled to 8\,kHz, converted to mel-spectrograms using 64 mel bins, an FFT size of 512, and hop length 64, normalized to $[0,1]$, and resized to $36\times36\times1$.
For Sign Language MNIST, the first ten available letter classes are remapped densely to indices $0$--$9$.

\subsection{Private Encoders and Dataset-Specific Decoders}
\label{sup:private_modules}

Each dataset first learns a private encoder--decoder pair $(E_m,D_m)$.
The private encoder maps an observation to a 128-dimensional private latent
vector $h^{(m)}$. The decoder receives this vector and a learned
64-dimensional embedding of the local class index:
\begin{equation}
\hat{x}^{(m)}=D_m(h^{(m)},k).
\end{equation}
The private reconstruction objective is
\begin{equation}
\mathcal{L}_{\mathrm{AE}}
=
10\|x-\hat{x}\|_1
+
5\left(1-\mathrm{SSIM}(x,\hat{x})\right)
+
2\|x-\hat{x}\|_2^2.
\label{sup:eq:ae_loss}
\end{equation}
Architectural details are summarized in
Table~\ref{sup:tab:architectures}. After this stage, the private encoders are
discarded and all $D_m$ are frozen.

\subsection{Unified Encoder and Address-Space Objective}
\label{sup:unified_encoder}

The unified encoder accepts variable spatial dimensions and produces
\begin{equation}
U(x)=\big(Z(x),q(x)\big),
\qquad
Z(x)\in\mathbb{R}^{10\times128}.
\end{equation}
For local index $k$, the selected address representation is $Z_k(x)$.
The auxiliary logits $q(x)$ are used only in the training loss.

For source dataset $m$, self-reconstruction is
\begin{equation}
\begin{aligned}
\mathcal{L}_{\mathrm{self}}^{(m)}
={}&
20\left\|x^{(m)}-D_m(Z_k(x^{(m)}),k)\right\|_1 \\
&+
10\left[1-\mathrm{SSIM}\!\left(x^{(m)},D_m(Z_k(x^{(m)}),k)\right)\right].
\end{aligned}
\label{sup:eq:self_loss}
\end{equation}
The auxiliary index loss is
\begin{equation}
\mathcal{L}_{\mathrm{idx}}
=
\mathrm{CE}\!\left(q(x^{(m)}),k\right).
\label{sup:eq:index_loss}
\end{equation}
For target dataset $n\neq m$, cross-module reconstruction is
\begin{equation}
\begin{aligned}
\mathcal{L}_{\mathrm{cross}}^{(m\rightarrow n)}
={}&
5\left\|x_k^{(n)}-D_n(Z_k(x^{(m)}),k)\right\|_1 \\
&+
3\left[1-\mathrm{SSIM}\!\left(x_k^{(n)},D_n(Z_k(x^{(m)}),k)\right)\right],
\end{aligned}
\label{sup:eq:cross_loss}
\end{equation}
where $x_k^{(n)}$ is an independently sampled observation from local index $k$ in dataset $n$.
Two of the remaining three datasets are sampled in each update.
No instance correspondence is used.

\subsection{Label-Free Receiver and Judge Protocol}
\label{sup:base_eval_protocol}

The unified encoder is frozen.
For each dataset, a label-free receiver is trained on selected address representations $Z_k(x)$ and receives no local index.
An independent judge is trained on real samples from the same dataset.
Receivers and judges use disjoint train, validation, and test splits.
The reported test split is unseen by both receiver and judge training, but is not guaranteed to be unseen by the original unified encoder.
For FSDD, real clips are split before augmentation; augmentation is applied only to the training split.

Address-representation substitution replaces the original address
representation with a real representation sampled from another address. We
report ARA under the original representation, ARA for the substituted target
address, retention of the original address, and zero/random controls.

\section{Extended SVHN Attachment Method}
\label{sup:svhn_method}

\subsection{SVHN Latent Space}
\label{sup:svhn_latent}

SVHN uses a separate ten-address encoder and a label-free decoder
\begin{equation}
D_{\mathrm{SVHN}}:
\mathbb{R}^{128}
\rightarrow
[0,1]^{32\times32\times3}.
\end{equation}
During SVHN latent-space training, the known local digit index selects the
corresponding SVHN latent representation.
The objective is
\begin{equation}
\mathcal{L}_{\mathrm{SVHN}}
=
8\|x-\hat{x}\|_1
+
5\left(1-\mathrm{SSIM}(x,\hat{x})\right)
+
2\|x-\hat{x}\|_2^2
+
\mathcal{L}_{\mathrm{cls}}.
\label{sup:eq:svhn_ae}
\end{equation}
For each digit $k$, we estimate the latent mean $\mu_k$ and standard deviation $\sigma_k$.

\subsection{Write Adapter}
\label{sup:write_adapter}

The write adapter maps a frozen address prototype and private noise into the SVHN latent space:
\begin{equation}
\hat z_{\mathrm{SVHN}}=W(p_k,\epsilon),
\qquad
\epsilon\sim\mathcal{N}(0,I).
\end{equation}
A diagonal class-conditional target is sampled as
\begin{equation}
\tilde z_k=\mu_k+\sigma_k\odot\epsilon.
\end{equation}
The write objective is
\begin{equation}
\mathcal{L}_{W}
=
\mathcal{L}_{\mathrm{cls}}^{W}
+
0.15\mathcal{L}_{\mathrm{lat}}^{W}
+
0.10\mathcal{L}_{\mathrm{div}}^{W},
\label{sup:eq:write_loss}
\end{equation}
where
\begin{equation}
\mathcal{L}_{\mathrm{cls}}^{W}
=
\mathrm{CE}
\left(
J_{\mathrm{train}}
\left(
D_{\mathrm{SVHN}}(W(p_k,\epsilon))
\right),k
\right),
\end{equation}
\begin{equation}
\mathcal{L}_{\mathrm{lat}}^{W}
=
\left\|W(p_k,\epsilon)-\tilde z_k\right\|_2^2,
\end{equation}
and
\begin{equation}
\mathcal{L}_{\mathrm{div}}^{W}
=
\left\|
|\hat z_1-\hat z_2|
-
|\tilde z_1-\tilde z_2|
\right\|_2^2.
\end{equation}
Separate SVHN judges are used for adapter optimization and final evaluation.

\subsection{Read Adapter and Transfer Metrics}
\label{sup:read_adapter}

The read adapter predicts one of the ten addresses:
\begin{equation}
R:\mathbb{R}^{128}\rightarrow\mathbb{R}^{10},
\qquad
\hat{k}=\arg\max R(z_{\mathrm{SVHN}}).
\end{equation}
The selected address prototype $p_{\hat{k}}$ is passed to the label-free MNIST receiver.

Address Routing Accuracy is
\begin{equation}
\mathrm{Acc}_{\mathrm{route}}
=
\frac{1}{N}
\sum_i
\mathbb{I}[\hat{k}_i=k_i].
\end{equation}
Target-side Address Realization Accuracy (ARA) is
\begin{equation}
\mathrm{ARA}
=
\frac{1}{N}
\sum_i
\mathbb{I}[J(D(p_{k_i}))=k_i].
\end{equation}
End-to-end Address Transfer Accuracy requires preservation of the correct address through the complete source-to-address-to-target path.
Transfer Retention is the ratio of End-to-end Address Transfer Accuracy to Address Routing Accuracy.
Interventions use the correct address prototype, a permuted address prototype, a zero vector, and a random latent vector.

\section{Extended CIFAR-100 Address-Code Method}
\label{sup:cifar_method}

CIFAR-100 provides 100 fine-class identities and 20 superclasses. Arithmetic and hierarchical codes are defined in Eqs.~\eqref{eq:arith_code} and \eqref{eq:hier_code} of the main paper. The experiment evaluates address-code reuse and does not perform CIFAR-100 image generation or semantic decomposition.

For a code
\begin{equation}
\phi(c)=[k_1,\ldots,k_L],
\end{equation}
each requested address $k_j$ selects the corresponding frozen address prototype $p_{k_j}$. The address prototype is decoded by a label-free receiver, and an independent judge predicts the realized address $\hat{k}_{c,j}$.

Target-side Address Realization Accuracy (ARA) is
\begin{equation}
\mathrm{ARA}
=
\frac{
\sum_c\sum_{j=1}^{L}
\mathbb{I}[\hat{k}_{c,j}=k_{c,j}]
}{
\sum_c L
},
\label{sup:eq:address_acc}
\end{equation}
and Class-Code Realization Accuracy (CRA) is
\begin{equation}
\mathrm{CRA}
=
\frac{1}{C}
\sum_{c=1}^{C}
\mathbb{I}[\hat{\phi}(c)=\phi(c)].
\label{sup:eq:code_acc}
\end{equation}
We separately report Round-trip Recovery Accuracy, Collision Count, and
Superclass Prefix Consistency.

\section{Extended Results}
\label{sup:results}

\subsection{Base Address-Space Intervention}
\label{sup:base_intervention}

\begin{table}[t]
\centering
\fontsize{7}{8}\selectfont
\caption{
Address-representation substitution results. All columns report ARA; chance
is $0.10$.
}
\label{sup:tab:base_intervention}
\setlength{\tabcolsep}{2.3pt}
\renewcommand{\arraystretch}{1.05}
\begin{tabular}{@{}lccccc@{}}
\toprule
\textbf{Dataset}
& \textbf{Original}
& \textbf{Perm.$\rightarrow$Target}
& \textbf{Perm.$\rightarrow$Original}
& \textbf{Zero}
& \textbf{Random} \\
\midrule
MNIST               & 0.986 & 0.985 & 0.001 & 0.098 & 0.099 \\
Fashion-MNIST       & 0.934 & 0.933 & 0.002 & 0.100 & 0.099 \\
FSDD                & 0.907 & 0.896 & 0.004 & 0.100 & 0.109 \\
Sign Language MNIST & 0.976 & 0.975 & 0.000 & 0.100 & 0.106 \\
\midrule
\textbf{Macro average}
                    & 0.951 & 0.947 & 0.002 & 0.100 & 0.103 \\
\bottomrule
\end{tabular}
\end{table}

This intervention run uses the substitution-evaluation sample set and is reported separately from the base receiver test in Table~\ref{tab:label_free_representation}; therefore, its original-representation macro ARA of 0.951 is not the base ARA of
0.9468. The original address almost disappears after substitution, while zero
and random representations remain at chance. Together with the label-free
receiver evaluation, these interventions confirm that the realized address is
controlled by the supplied continuous address representation rather than
a decoder-side address label.

\subsection{SVHN Transfer Decomposition and Intervention}
\label{sup:svhn_results}

\begin{table}[t]
\centering
\fontsize{7}{8}\selectfont
\caption{
Internal CIIR decomposition for bidirectional MNIST--SVHN transfer.
}
\label{sup:tab:svhn_decomposition}
\resizebox{\columnwidth}{!}{
\begin{tabular}{lcccc}
\toprule
\textbf{Direction}
& \textbf{Address Routing}
& \textbf{Target ARA}
& \textbf{End-to-end Transfer}
& \textbf{Transfer Retention} \\
\midrule
MNIST$\rightarrow$SVHN & 0.9095 & 0.9918 & 0.9024 & 0.9922 \\
SVHN$\rightarrow$MNIST & 0.8229 & 1.0000 & 0.8229 & 1.0000 \\
\bottomrule
\end{tabular}
}
\end{table}

\begin{table}[t]
\centering
\fontsize{7}{8}\selectfont
\caption{
MNIST$\rightarrow$SVHN address-space intervention.
Chance is $0.10$.
}
\label{sup:tab:svhn_intervention}
\resizebox{\columnwidth}{!}{
\begin{tabular}{ccccc}
\toprule
\textbf{Original}
& \textbf{Perm.$\rightarrow$Target}
& \textbf{Perm.$\rightarrow$Original}
& \textbf{Zero}
& \textbf{Random} \\
\midrule
0.9918 & 0.9902 & 0.0034 & 0.0958 & 0.0928 \\
\bottomrule
\end{tabular}
}
\end{table}

The forward path is limited by MNIST-to-address routing, whereas the reverse
path is limited by SVHN-to-address routing. Once the correct address is available, target-side ARA is $99.18\%$ for
SVHN and $100\%$ for MNIST. Permuting the address prototype redirects the
output to the substituted address with $99.02\%$ ARA.  Secondary SVHN generation diagnostics were SSIM $=0.8017$ and FID $=62.50$, with a real--real FID reference of $19.66$. These metrics describe output quality but are not the primary evidence for address transfer.

\subsection{CIFAR-100 Address-Code Analysis}
\label{sup:cifar_results}

\begin{table}[t]
\centering
\fontsize{7}{8}\selectfont
\caption{
Direct arithmetic address-code realization using the frozen MNIST-derived
address prototypes. $k_1$ ARA and $k_2$ ARA report realization at each code
position; ARA averages both positions, and CRA requires both to be correct.
}
\label{sup:tab:cifar_direct}
\resizebox{\columnwidth}{!}{
\begin{tabular}{lcccc}
\toprule
\textbf{Receiver}
& \textbf{$k_1$ ARA}
& \textbf{$k_2$ ARA}
& \textbf{ARA}
& \textbf{CRA} \\
\midrule
MNIST
& 1.0000 & 1.0000 & 1.0000 & 1.0000 \\
Fashion-MNIST
& 0.8000 & 0.8000 & 0.8000 & 0.6400 \\
FSDD
& 0.4000 & 0.4000 & 0.4000 & 0.1600 \\
Sign Language MNIST
& 0.1000 & 0.1000 & 0.1000 & 0.0100 \\
\bottomrule
\end{tabular}
}
\end{table}

The arithmetic and hierarchical assignments both achieved 100\% Round-trip Recovery Accuracy with zero collisions. The hierarchical assignment additionally achieved 100\% Superclass Prefix Consistency.

\begin{figure*}[t]
\centering
\fontsize{7}{8}\selectfont
\includegraphics[height=0.75\textheight,keepaspectratio]{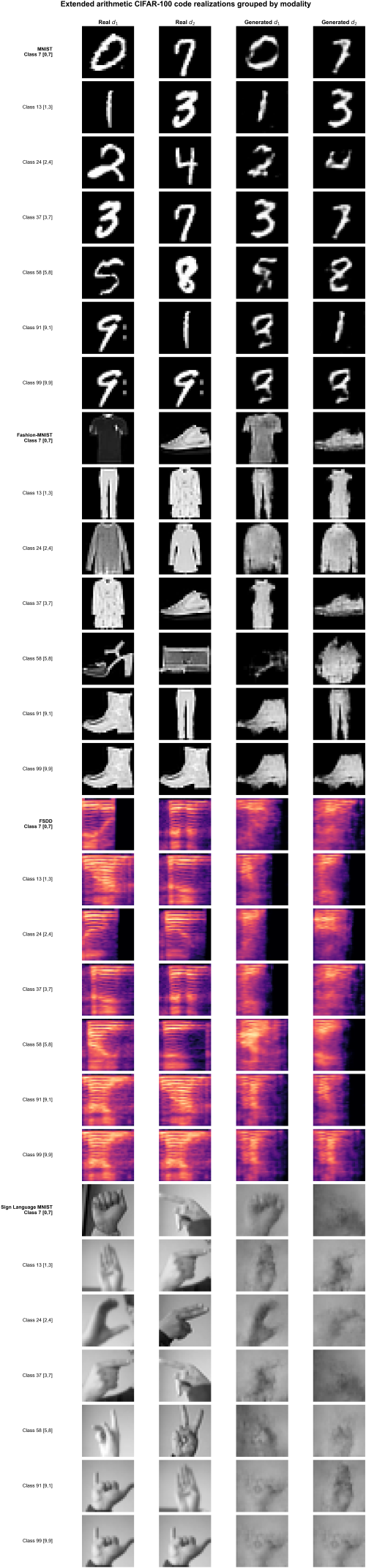}
\caption{
Extended arithmetic two-address code realizations grouped by modality. Each example maps a known CIFAR-100 fine-class identity $c$ to the ordered two-address decimal code $\phi_{\mathrm{arith}}(c)=[\lfloor c/10\rfloor,c\bmod 10]$. For every modality and target class, the first two columns show real examples associated with the two addresses, while the final two columns show their generated realizations through the corresponding frozen label-conditioned decoder. The figure visualizes reuse of the ten addresses as an ordered class-code vocabulary rather than CIFAR-100 image generation. This qualitative decoder visualization is distinct from the label-free receiver evaluation reported in Table~\ref{sup:tab:cifar_direct}.
}
\label{fig:supp_cifar_arithmetic_extended}

\end{figure*}

\begin{figure*}[t]
\centering
\fontsize{7}{8}\selectfont
\includegraphics[height=0.75\textheight,keepaspectratio]{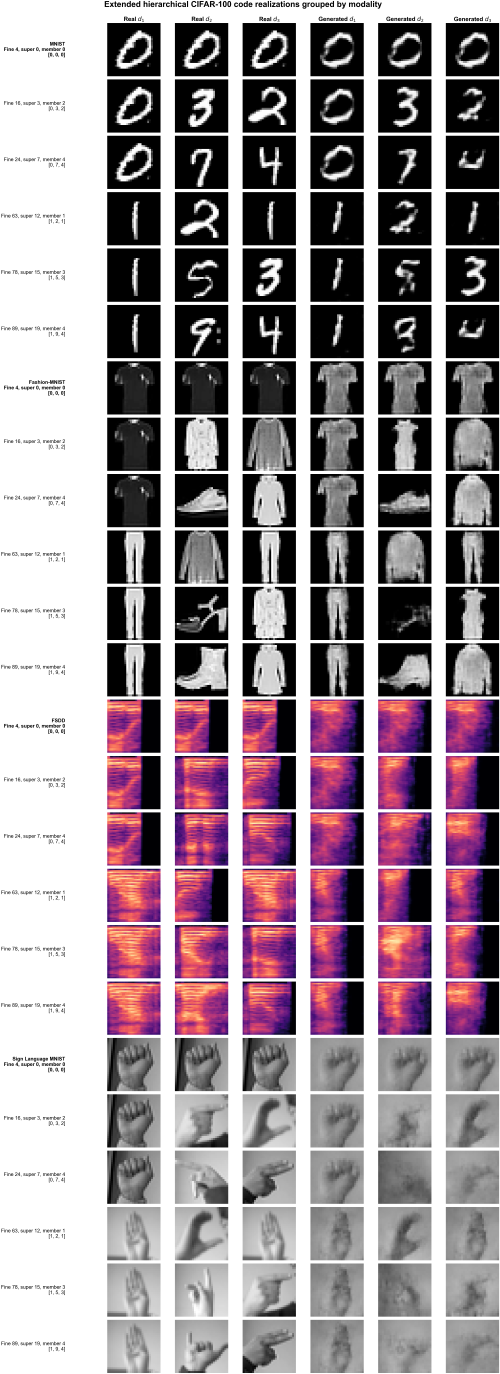}
\caption{
Extended hierarchical three-address code realizations grouped by modality. For a CIFAR-100 fine class $c$, the code $\phi_{\mathrm{hier}}(c)= [\lfloor s(c)/10\rfloor,s(c)\bmod 10,m(c)]$ represents its superclass $s(c)$ using the first two addresses and its within-superclass member address $m(c)$ using the third. For each modality, the first three columns show real examples associated with the requested addresses and the final three columns show their generated realizations. Examples belonging to the same superclass preserve a common two-address prefix, illustrating hierarchical organization through ordered address reuse. This qualitative decoder visualization illustrates the deterministic code structure; quantitative hierarchical execution was not evaluated.
}
\label{fig:supp_cifar_hierarchical_extended}

\end{figure*}

Direct arithmetic-code execution was exact through the MNIST receiver. The Fashion-MNIST receiver realized most requested addresses, whereas FSDD and Sign Language MNIST were less compatible with the MNIST-derived address-prototype bank. These differences reflect receiver compatibility rather than ambiguity in the deterministic address codes. Extended arithmetic and hierarchical realizations are shown in Figs.~\ref{fig:supp_cifar_arithmetic_extended} and~\ref{fig:supp_cifar_hierarchical_extended}.

\subsection{Exploratory CIFAR-Conditioned Address-Representation Generation}
\label{sup:cifar_expansion}

We additionally tested a stronger task in which a class-conditioned expansion encoder receives a grayscale CIFAR-100 image and its known fine-class identity and attempts to generate the two continuous address representations required by the arithmetic code. This experiment is distinct from direct code execution and is not required to establish the collision-free class coding result.

\begin{table}[!h]
\centering
\fontsize{7}{8}\selectfont
\caption{
Exploratory CIFAR-conditioned generation of arithmetic-code address
representations. $k_1$ ARA and $k_2$ ARA report each code position; ARA
averages both positions, and CRA requires both to be correct.
}
\label{sup:tab:cifar_expansion}
\resizebox{\columnwidth}{!}{
\begin{tabular}{lcccc}
\toprule
\textbf{Receiver}
& \textbf{$k_1$ ARA}
& \textbf{$k_2$ ARA}
& \textbf{ARA}
& \textbf{CRA} \\
\midrule
MNIST
& 0.2699 & 0.1599 & 0.2149 & 0.0400 \\
Fashion-MNIST
& 0.5086 & 0.4000 & 0.4543 & 0.2000 \\
FSDD
& 0.0000 & 0.2035 & 0.1018 & 0.0000 \\
Sign Language MNIST
& 0.3000 & 0.2371 & 0.2686 & 0.0900 \\
\bottomrule
\end{tabular}
}
\end{table}

The expansion encoder did not reliably generate the requested address representations. Its best result was obtained through the Fashion-MNIST receiver, with 45.4\% ARA and 20.0\% CRA. This result identifies CIFAR-conditioned address-representation generation as a harder extension beyond direct execution of the predefined address codes.

\subsection{Scope of the Address-Code Claim}
\label{sup:cifar_scope}

The CIFAR-100 experiment evaluates address-code reuse.
A code such as $[5,2]$ identifies class 52 but does not assert that the class is semantically composed of address meanings 5 and 2. The experiment models a new dataset whose class ontology is known, while exhaustive cross-modal pairing is unavailable or expensive. CIIR reuses a fixed address vocabulary rather than introducing one new address for every class. % ============================================================
% SUPPLEMENT FIGURE S1 — EXTENDED TEN-INDEX REALIZATION
% ============================================================
\begin{figure*}[t]
\centering
\fontsize{7}{8}\selectfont
\includegraphics[width=0.5\textwidth]{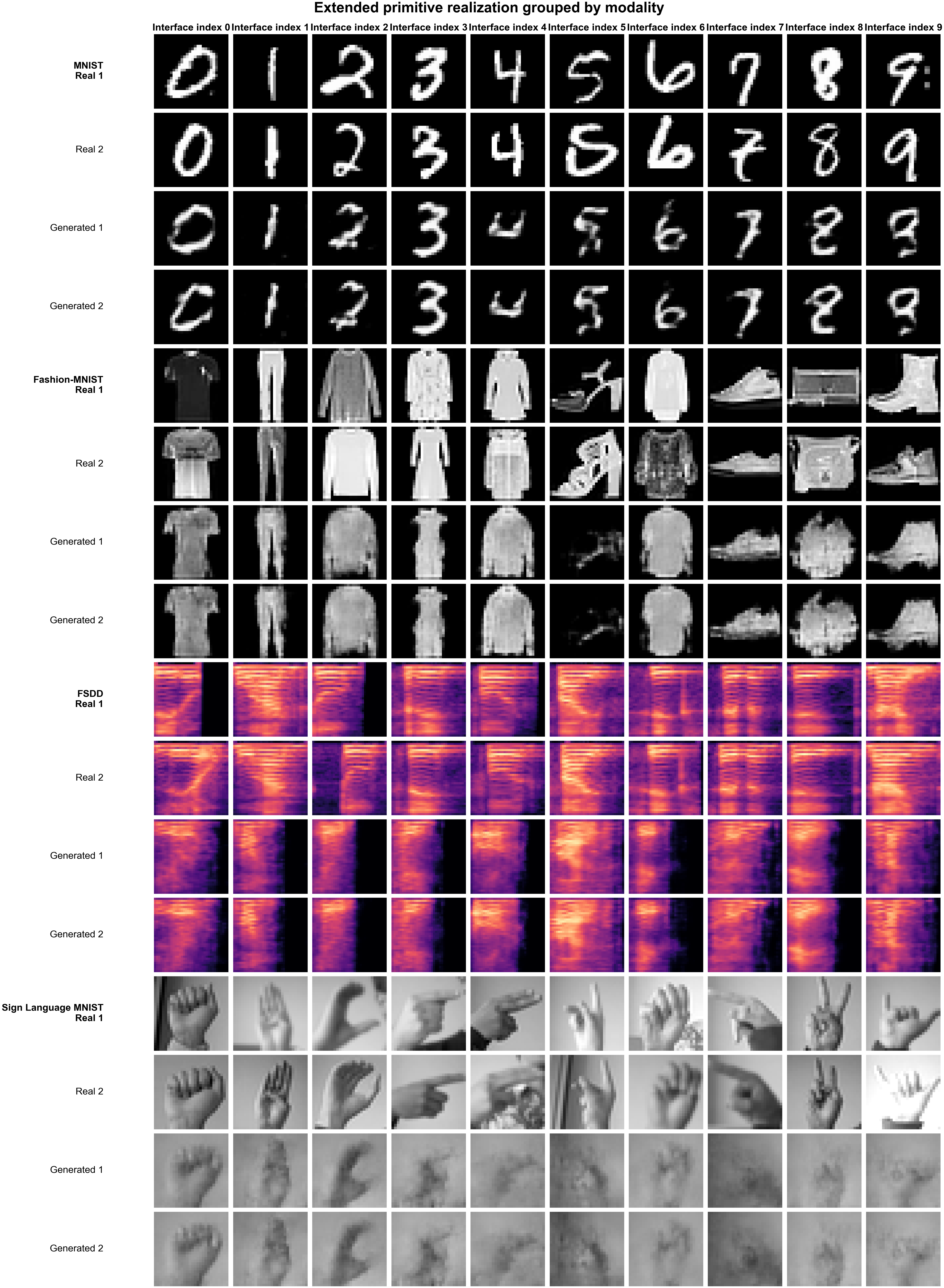}
\caption{
Extended qualitative comparison of the ten CIIR addresses, grouped by output modality. For each modality and local index, two real examples and two generated examples are shown. Generated outputs use the two actual MNIST address representations nearest to the corresponding address centroid, together with the requested local index, through the original frozen dataset-specific decoder. The use of address-centroid-nearest representations provides a deterministic representative selection procedure rather than manual visual selection.
}
\label{fig:supp_base10_extended}

\end{figure*}

% ============================================================
% SUPPLEMENT FIGURE S2 — EXTENDED ARITHMETIC CODES
% ============================================================

% ============================================================
% SUPPLEMENT FIGURE S3 — EXTENDED HIERARCHICAL CODES
% ============================================================

\end{document}